\pdfoutput=1

\documentclass[11pt]{article}

\usepackage[final]{acl}

\usepackage{times}
\usepackage{latexsym}

\usepackage[T1]{fontenc}

\usepackage[utf8]{inputenc}

\usepackage{microtype}

\usepackage{inconsolata}

\usepackage{graphicx}
\usepackage{booktabs}
\usepackage{amsmath}
\usepackage{amssymb}
\usepackage{tcolorbox}
\usepackage{caption} 
\usepackage{cleveref} 
\usepackage{bm}
\usepackage{multirow}

\title{IRT-Router: Effective and Interpretable Multi-LLM Routing \\via Item Response Theory}

\author{Wei Song$^1$, Zhenya Huang$^{1,2}$\thanks{Corresponding author}, Cheng Cheng$^1$, Weibo Gao$^1$,  Bihan Xu$^1$, \\ {\bf Guanhao Zhao$^{1}$, Fei Wang$^{3}$,  Runze Wu$^{4}$}\\
\normalsize$^1$State Key Laboratory of Cognitive Intelligence, University of Science and Technology of China \\ 
\normalsize$^2$Institute of Artificial Intelligence, Hefei Comprehensive National Science Center\\
\normalsize$^3$School of Computing, National University of Singapore\\
\normalsize$^4$NetEase Fuxi AI Lab\\
\small \texttt{\{sw2, doublecheng, weibogao, xbh0720, ghzhao0223\}@mail.ustc.edu.cn}, \small \texttt{huangzhy@ustc.edu.cn}, \\ \small \texttt{wang-fei@nus.edu.sg},  \small \texttt{wurunze1@corp.netease.com}\\ 
}

\begin{document}
\maketitle
\begin{abstract}
Large language models (LLMs) have demonstrated exceptional performance across a wide range of natural language tasks. However, selecting the optimal LLM to respond to a user query often necessitates a delicate balance between performance and cost. While powerful models deliver better results, they come at a high cost, whereas smaller models are more cost-effective but less capable. To address this trade-off, we propose IRT-Router, a multi-LLM routing framework that efficiently routes user queries to the most suitable LLM. Inspired by Item Response Theory (IRT), a psychological measurement methodology, IRT-Router explicitly models the relationship between LLM capabilities and user query attributes. This not only enables accurate prediction of response performance but also provides interpretable insights, such as LLM abilities and query difficulty. Additionally, we design an online query warm-up technique based on semantic similarity, further enhancing the online generalization capability of IRT-Router. Extensive experiments on 20 LLMs and 12 datasets demonstrate that IRT-Router outperforms most baseline methods in terms of effectiveness and interpretability. Its superior performance in cold-start scenarios further confirms the reliability and practicality of IRT-Router in real-world applications. Code is available at \url{https://github.com/Mercidaiha/IRT-Router}.
\end{abstract}

\section{Introduction}
\label{intro}
\begin{figure}[t]
  \centering
  \includegraphics[width=\columnwidth]{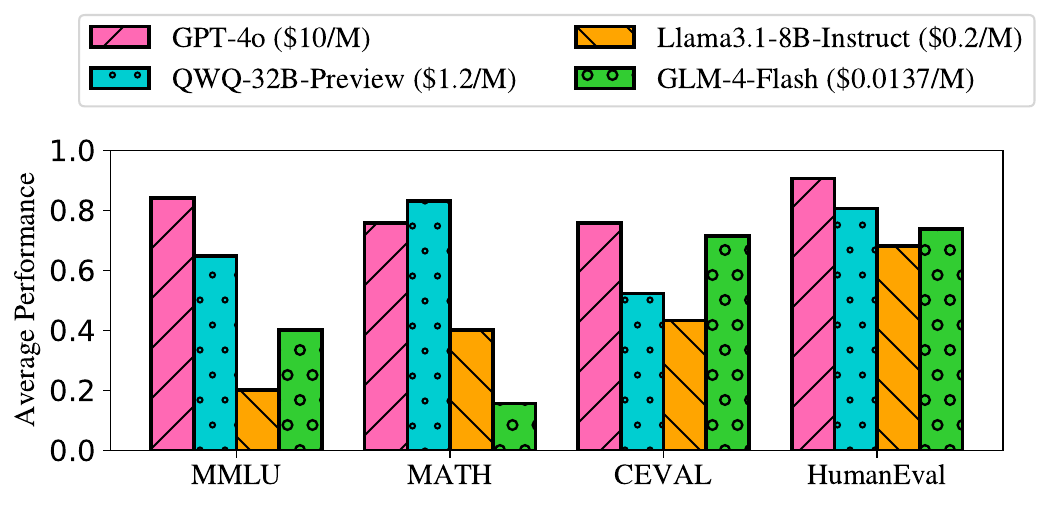}
  \caption{Four representative LLMs' output pricing and their performance on four different datasets.}
  \label{fig:diff}
\end{figure}

In recent years, large language models (LLMs) have demonstrated exceptional capabilities across a wide range of natural language tasks \cite{liu2024deepseek,yang2024qwen2,liu2024socraticlm,zhao2024comprehensive,xue2024decompose}, rapidly becoming a dominant force in the field of natural language processing. Generative applications based on LLMs, such as ChatGPT, have attracted widespread usage from various industries due to their outstanding accessibility. Users can input queries in natural language and receive responses without the need for specialized data or code. As user demands become increasingly complex, new LLMs are released almost daily. These models vary significantly in terms of reasoning ability, performance, computational resource requirements, and cost, as shown in Figure~\ref{fig:diff}. Generally, larger models tend to provide stronger performance but come with higher computational costs, while smaller models, though more affordable, often exhibit weaker performance.

This diverse LLM ecosystem presents a dilemma for practical applications: How can user queries be effectively routed to the most appropriate LLM? While routing all queries to the largest and most powerful model ensures high-quality results, this approach is costly and unnecessary. For simpler queries, smaller models are often sufficient, offering cheaper and faster solutions. On the other hand, using large models like o1 or enhanced LLMs through complex strategies may result in ``overthinking'' \citep{chen2024think23overthinkingo1like}, leading to significant resource waste and even reducing answer quality \citep{jeong2024adaptive, xu2024adaption, ma2025debate}. Therefore, finding the optimal balance between response quality and cost has become a key challenge for the practical application of LLMs.

\begin{figure}[]
\centering
  \includegraphics[width=\columnwidth]{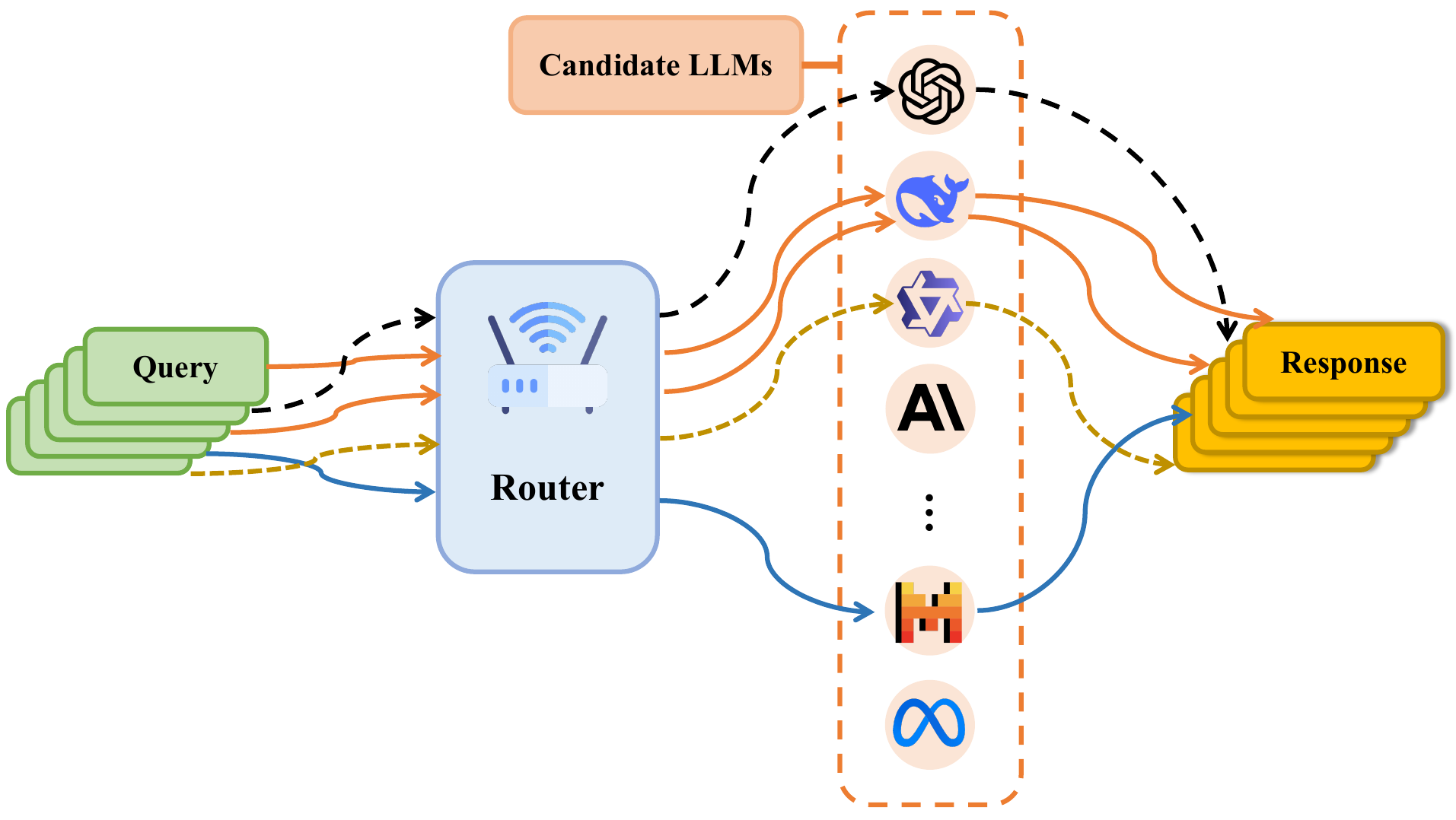}
  \caption{LLM Router. Queries are assigned to different LLMs for responses through the trained router.}
  \label{fig:router}
\end{figure}

LLM Router offers an efficient solution by assigning each user query to the most appropriate LLM through a custom router (see Figure \ref{fig:router}). This approach aims to maximize response performance within cost constraints or minimize cost while maintaining target quality. Early LLM routers used static strategies, routing queries to progressively costlier models until the desired quality was achieved \cite{chen2023frugalgpt}, resulting in resource wastage from ineffective requests. Recent data-driven routing methods use pre-trained response performance predictors to predict the response quality from each LLM after receiving a query and routing the query to the optimal LLM \cite{ding2024hybrid,hu2024routerbench}. Due to their low routing costs and high efficiency, data-driven routing has become the mainstream approach in LLM router research.

However, existing data-driven routing methods still face limitations in \textbf{effectiveness} and \textbf{interpretability}: (1) In terms of effectiveness, current approaches often rely on existing models like BERT to predict LLM response quality. However, they lack a systematic and rational framework tailored to the LLM router field, which hinders their ability to fully exploit the underlying relationships between LLMs and queries.
Furthermore, the randomness and openness of user queries cause discrepancies between queries in the online environment and those during the training phase, leading to a cold-start problem that further limits the effectiveness of LLM router. (2) In terms of interpretability, current methods only output performance scores of LLM responses, without providing the rationale behind the predictions. Providing a clear explanation, such as ``Routing the math query to \textit{QwQ-32B-Preview} because it performs better on math problem'' (as shown in Figure~\ref{fig:diff}), would significantly enhance the reliability of predictions. Transparent explanations not only help users understand the decision-making process of the model but also increase trust and acceptance.

To address these challenges, we propose \textbf{IRT-Router}, a multi-LLM routing framework built on Item Response Theory (IRT). IRT is a psychological theory widely used to measure human test-takers' abilities, which assumes that test-takers have a ``latent ability'' to answer questions, while questions possess attributes such as ``latent difficulty.'' By modeling the probability of test-takers providing correct answers, IRT explicitly captures the implicit relationship between human abilities and test item attributes.
In our context, each LLM is treated as a ``test-taker'' with a latent, multidimensional ability to respond to user queries, while each query is treated as a ``question'' with latent attributes. Based on IRT, we explicitly model the relationship between the multidimensional abilities of LLMs and query attributes (e.g., response difficulty) to predict the performance of each LLM on specific queries. Combined with cost optimization objectives (e.g., output pricing), IRT-Router selects the most suitable LLM for response.
Compared to existing methods, IRT-Router is specifically designed for this domain, grounded in psychological measurement theory, making its modeling more principled. Additionally, IRT-Router quantifies attributes such as the latent abilities of LLMs and the response difficulty of queries, providing interpretable justifications for routing decisions. To mitigate the cold-start problem for new queries after online deployment, we warm up new queries using semantically similar existing queries. In particular, we design our model based on two concrete implementations of IRT families, namely MIRT-Router based on Multidimensional IRT~\cite{reckase2009multidimensional} and NIRT-Router based on NCDM~\cite{wang2020neural}. Extensive experiments validate the effectiveness and interpretability of our approach.

The main contributions of this work are summarized as follows:

\raisebox{0.25ex}{\small{$\bullet$}}\normalsize\; We innovatively apply psychological measurement theory to the LLM routing field, exploring a rational way to combine data mining techniques with LLM routing tasks.

\raisebox{0.25ex}{\small{$\bullet$}}\normalsize\; IRT-Router is a novel framework tailored to LLM routing. It can explicitly establish the relationship between query attributes and LLM ability, ensuring both effectiveness and interpretability.

\raisebox{0.25ex}{\small{$\bullet$}}\normalsize\; Extensive experiments with 20 LLMs and 12 datasets demonstrate that IRT-Router outperforms most baseline methods in terms of effectiveness and interpretability. Its superiority in cold-start scenarios confirms that IRT-Router is more realistic and reliable for practical applications.

\section{Related Work}
\paragraph{Item Response Theory.} Item Response Theory (IRT) \cite{woodruff1996estimation} is a widely used psychological theory for measuring human test-takers' abilities. It assumes that test-takers possess a ``latent ability'' to answer questions, while the questions themselves have attributes, such as ``latent difficulty''. By modeling the probability that test-takers provide correct answers, IRT explicitly captures implicit relationship between human abilities and the attributes of questions. Specially, IRT relies on the psychological \textbf{Monotonicity} assumption, which states that \textit{the probability of a test-taker answering a question correctly is proportional to their proficiency in the skill associated with the item}, thus ensuring interpretability.

In machine learning, IRT has been implemented in various forms to diagnose human abilities by modeling response data \cite{gao2021rcd, gao2023leveraging,li2025foundation,zhang2023fairlisa,zhang2024towards,liu2019ekt}. For instance, IRT and MIRT models \cite{woodruff1996estimation, reckase2009multidimensional} use logistic-like functions to model unidimensional and multidimensional learner abilities, respectively. Meanwhile, the NCDM model \cite{wang2020neural} leverages neural networks to capture higher-order interactions between learners and test items, enabling the evaluation of multidimensional abilities. Recently, due to its effectiveness and interpretability in human measurement, IRT has been applied to assess machine learning models \cite{liu2024multi, martinez2019item}, evaluate sample difficulty \cite{martinez2022ai}, enhance recommendation systems \cite{liu2023we}, rank leaderboards \cite{rodriguez2021evaluation}, and evaluate LLM capabilities \cite{guinet2024automated, gor2024great, liu2024leveraging}.

We are motivated to model the relationship between LLM and query based on IRT to enhance the effectiveness and interpretability of LLM router.

\paragraph{LLM Router.}
The LLM Router field remains in an exploratory phase \cite{vsakota2024fly, liu2024optllm, ramirez2024optimising, hari2023tryage, mohammadshahi2024routoolearningroutelarge, dai2024cost, stojkovic2024dynamollm}. Unlike Mixture of Experts (MoE)\cite{cai2024survey}, which requires loading all expert parameters on a single machine, or LLM ensembling, which requires outputs from all candidate models, LLM routers only assign queries to the most suitable model, improving performance while reducing costs.

Earlier approaches like FrugalGPT and AutoMix \cite{chen2023frugalgpt, aggarwal2023automix} cascade queries through models ordered by cost, obtaining responses until one is deemed sufficient. Other methods use data-driven techniques to train lightweight routers for optimal LLM assignment.
HybridLLM \cite{ding2024hybrid} uses a BERT-based router, while RouteLLM \cite{ong2024routellm} introduces routers like SW-ranking, MF, and BERT classifiers, focusing mainly on binary routing (large vs. small models). Zooter \cite{lu2023routing} and RouterDC \cite{chen2024routerdc} leverage smaller models with reward mechanisms or contrastive learning, achieving performance similar to larger models. GraphRouter \cite{feng2024graphrouter} uses a GNN-based router but requires prior task knowledge, which can be challenging in real-world use. RouterBench \cite{hu2024routerbench} and \citep{shnitzer2023large} propose KNN-based routing.

LLM routers are crucial in commercial systems that reduce costs for LLM applications, such as Martian (\textit{withmartian.com}) and Neutrino AI (\textit{neutrinoapp.com}). Martian claims to ``beat GPT-4 on performance and reduce costs by 20\%-97\%'' through dynamic model routing. LLM API providers like OpenRouter (\textit{openrouter.ai}) also offer similar capabilities.

\begin{figure*}[]
  \centering
  \includegraphics[width=\textwidth]{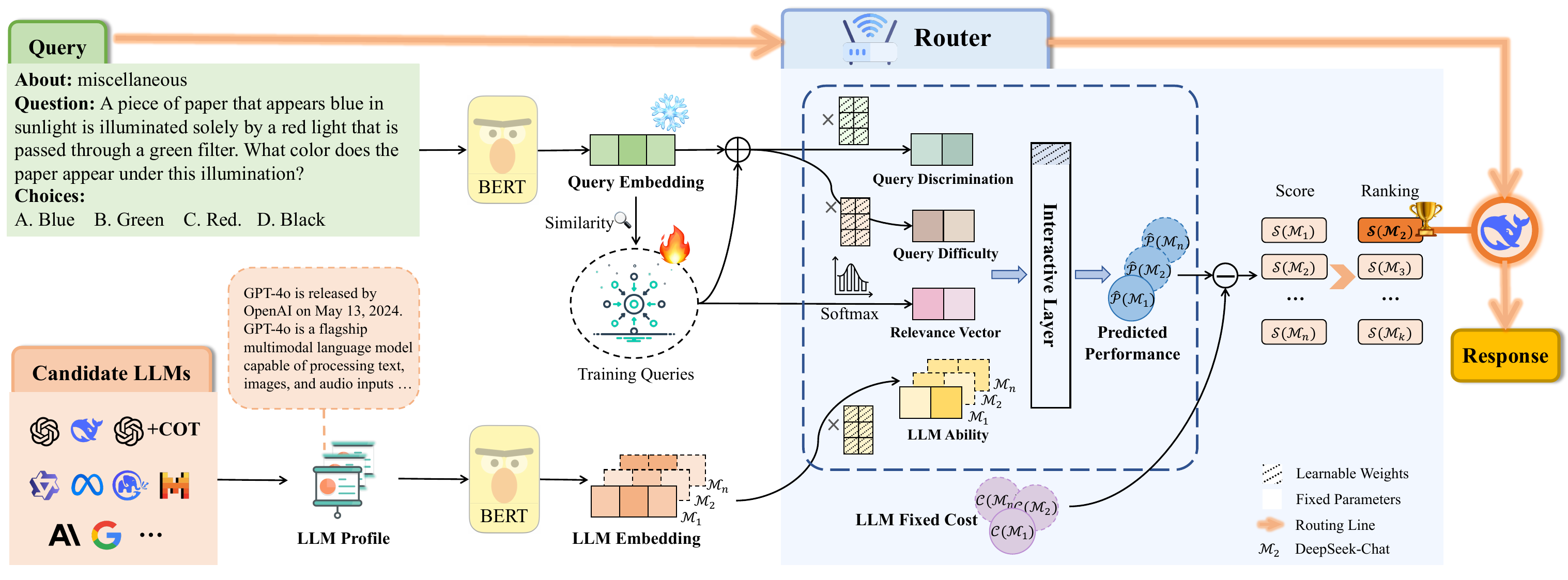}
  \caption{Framework of IRT-Router. The left side represents query and LLM embedding, the middle performs IRT-based prediction, and the right side outputs the routing decision.}
  \label{fig:frame}
\end{figure*}

\section{Preliminary}
\subsection{Problem Definition}
\label{pd}
Given a set of large language models (LLMs) $\mathcal{M} = \{ M_1, M_2, \dots, M_n \}$ and a set of queries $\mathcal{Q} = \{ q_1, q_2, \dots, q_m \}$, our goal is to assign each query $q_i \in \mathcal{Q}$ to the most suitable LLM $M_j \in \mathcal{M}$, achieving \textbf{higher} performance and \textbf{lower} cost.

For each query $q_i$, we define a scoring function:
\begin{equation}
    \mathcal{S}(q_i, M_j) = \alpha \cdot \hat{\mathcal{P}}(q_i, M_j) - \beta \cdot \mathcal{C}(M_j),
\label{score}
\end{equation}
where:



\noindent\raisebox{0.25ex}{\small{$\bullet$}}\normalsize\; $\hat{\mathcal{P}}$ is the predicted performance of model $M_j$ on query $q_i$, obtained from a trained model.

\noindent\raisebox{0.25ex}{\small{$\bullet$}}\normalsize\; $\mathcal{C}(M_j)$ represents the fixed cost of using LLM $M_j$. Here, to unify the measurement, we define it as the linear mapping of LLM output pricing (see Table \ref{tab:llms}) to the range $[0,1]$\footnote{For example, the most expensive candidate LLM, GPT-4o's output pricing is \$10/M. So $\mathcal{C}$(GPT-4o) is $10/10=1$, while $\mathcal{C}$(DeepSeek-Chat) is $0.28/10=0.028$.}. Here, we approximate $\mathcal{C}(M_j)$ as a fixed cost just for simplicity, which is the most basic representation of cost. Actually, $\mathcal{C}(M_j)$ can be adjusted based on user-defined settings. For instance, if a user has ample computational resources, the cost for self-hosted open-source LLMs can even be set to 0.

\noindent\raisebox{0.25ex}{\small{$\bullet$}}\normalsize\; $\alpha$ and $\beta$ are predefined trade-off parameters controlling the relative importance of performance and cost, with $\alpha + \beta = 1$. A larger $\alpha$ indicates a higher emphasis on performance, whereas a larger $\beta$ prioritizes cost efficiency.

The optimal model assignment is determined by selecting the model that maximizes the score:
\begin{equation}
    M^*(q_i) = \arg\max_{M_j \in \mathcal{M}} \mathcal{S}(q_i, M_j).
\end{equation}

Thus, the key challenge lies in accurately learning the performance prediction function $\hat{\mathcal{P}}(q_i, M_j)$ to ensure effective query routing.

\subsection{Item Response Theory}
\label{irt}
IRT is a psychometric theory used to measure the ability of human test-takers based on their responses to test items \cite{woodruff1996estimation}. In the context of our work, the LLMs act as the test-takers, and the queries represent the items. For a given query \( q_i \) and a LLM \( M_j \), The predicted performance of \( M_j \) on \( q_i \) can be modeled as:
\begin{equation}
\hat{\mathcal{P}}(q_i, M_j) = IRT(\theta_{M_j};b_i, a_i, ...),
\label{eq:irt}
\end{equation}
where, \(IRT(\cdot)\) is a general form and has various implementations, such as logistic functions in the IRT model and MIRT model \cite{woodruff1996estimation,reckase2009multidimensional}, and neural networks in the NCDM model \cite{wang2020neural}. \textbf{For LLM modeling}, the key parameter is \(\theta_{M_j}\), representing the model's ability. \textbf{For query modeling}, there are more parameters. For instance, \(b_i\) is the difficulty parameter, which captures the inherent difficulty of query \(q_i\), and \(a_i\) is the discrimination parameter, which controls how sharply the predicted performance changes as \(\theta_{M_j}\) increases.

Specially, IRT relies on the psychological \textbf{Monotonicity} assumption, which states that as the LLM's ability \( \theta_{M_j} \) increases, the predicted performance on query increases. This assumption aligns with the intuitive idea that more capable LLMs are more likely to perform well on more difficult queries, ensuring the  interpretability of LLM router.

\section{Methods}
As shown in Figure \ref{fig:frame}, the proposed framework of IRT-Router operates as follows: Initially, we obtain the embeddings of both the query and the candidate LLMs. Next, the performance of each LLM is predicted using an IRT-based model. These performance predictions are then combined with the LLM’s fixed costs to compute ranking scores. Finally, the query is routed to the LLM with the highest score for generating the response.

\subsection{Query and LLM Embeddings}
Each query $q_i$ is first transformed into a query embedding $\mathbf{e}_{q_i} \in \mathbb{R}^{d_q}$ using a pre-trained embedding model (e.g., BERT). This embedding captures the semantic meaning of the query. 

Similarly, each LLM \( M_j \) (e.g., GPT-4o) is represented by a corresponding embedding \( \mathbf{e}_{M_j} \in \mathbb{R}^{d_M} \), which is derived from its profile. The profile includes metadata such as the model's release date, developer, type, key features, and a brief description (see Table \ref{tab:profile}). This profile is then encoded to form the LLM's embedding.
\subsection{IRT-based Prediction}
We have mentioned in Eq.(\ref{eq:irt}) that there are multiple implementation forms of IRT. Here, we will introduce a lightweight version, \textbf{M}ultidimensional\textbf{IRT}-Router, followed by a more interpretable version, \textbf{N}eural\textbf{IRT}-Router.
\subsubsection{MIRT-Router}
Inspired by Multidimensional Item Response Theory~\cite{reckase2009multidimensional}, MIRT-Router models the interaction between queries and LLMs using a logistic function, where each LLM is described by its latent ability $\boldsymbol{\theta}_{M_j} \in \mathbb{R}^\mathcal{N}$ in multiple dimensions, and each query is characterized by its discrimination $\mathbf{a}_i \in \mathbb{R}^\mathcal{N}$ and difficulty $b_i \in \mathbb{R}$. 
These parameters are all obtained through transformation layers: 
\begin{align}
    \boldsymbol{\theta}_{M_j} = \mathbf{W}_\theta \mathbf{e}_{M_j}, 
    \mathbf{a}_i = \mathbf{W}_a \mathbf{e}_{q_i},
    b_i = \mathbf{W}_b \mathbf{e}_{q_i},
\end{align}
where $\mathbf{W}_\theta$, $\mathbf{W}_a$ and $\mathbf{W}_b$ are all learnable weights.

\paragraph{Interactive Function.}
The predicted performance of $M_j$ on $q_i$ follows the logistic function:
\begin{equation}
    \hat{\mathcal{P}}(q_i, M_j) = \frac{1}{1 + \exp(-\mathbf{a}_i^\top \boldsymbol{\theta}_{M_j} + b_i)}.
\end{equation}
\paragraph{Training.}
\label{train}
We train MIRT-Router on a dataset $\mathcal{D}_{\text{train}}$ containing tuples $(q_i, M_j, y_{ij})$, where $y_{ij}$ is the empirical performance score of $M_j$ on $q_i$. This score is computed by comparing the LLM's response to the ground truth.

To learn the weights, we minimize the binary cross-entropy loss:
\begin{equation}
\begin{split}
    \mathcal{L} &= - \sum_{(q_i, M_j, y_{ij}) \in \mathcal{D}_{\text{train}}} \left[ y_{ij} \log \hat{\mathcal{P}}(q_i, M_j) \right. \\
    &\quad \left. + (1 - y_{ij}) \log (1 - \hat{\mathcal{P}}(q_i, M_j)) \right].
    \label{loss}
\end{split}
\end{equation}

\subsubsection{NIRT-Router}
While MIRT-Router focuses on latent abilities, NIRT-Router extends it by incorporating an explicit relevance vector $\mathbf{r}_{q_i}$ that associates each dimension with specific ability which is predefined (see Appendix \ref{appendix:abilities}).
\paragraph{Relevance Vector}
\label{rele}
The relevance vector \( \mathbf{r}_{q_i} \in \mathbb{R}^\mathcal{N} \) represents the degree to which a query \( q_i \) is associated with different ability dimensions. 

For \( \mathcal{Q}_{\text{train}} \): To define the relevance vector \(\mathbf{r}_{q_i}\), we first perform clustering on the question embeddings using UMAP \cite{mcinnes2018umap} for dimensionality reduction followed by HDBSCAN \cite{mcinnes2017hdbscan} clustering, which can adaptively identify clusters through density analysis. Each cluster represents a set of questions that share similar ability requirements. For each cluster, we identify the relevant abilities by considering the abilities required by $num (=5)$ sample questions within the cluster, which are done through a LLM for convenience (see Appendix \ref{appendix:prompt}). The relevance vector \(\mathbf{r}_{q_i}\) for a given query  \(q_i\)  is then constructed by assigning 1 or 0 to each ability dimension, indicating whether the ability is relevant to the query.

For \( \mathcal{Q}_{\text{test}} \): Since the true relevance vectors are not available, we approximate \( \mathbf{r}_{q_i} \) using the mean relevance vectors of its 5-nearest neighbors (5-NN) in the embedding space. This ensures that unseen queries still have reasonable relevance estimations.  

\paragraph{Interactive Function.}  
To predict performance of $M_j$ on $q_i$, NIRT-Router applies a neural interaction layer:  
\begin{align}
    \mathbf{x}_{ij} = \mathbf{r}_{q_i} \odot (\boldsymbol{\theta}_{M_j} - \mathbf{b}_i) \times a_i, \\
    \hat{\mathcal{P}}(q_i, M_j) = \sigma(\phi(\mathbf{W}_1 \mathbf{x}_{ij}^\top + \mathbf{b}_1)),
\end{align}
where:
\begin{equation}
\begin{split}
\boldsymbol{\theta}_{M_j} = \sigma(\mathbf{W}_\theta \mathbf{e}_{M_j}) \in \mathbb{R}^\mathcal{N},\\ a_i = \mathbf{W}_a  \mathbf{e}_{q_i} \in \mathbb{R},  \mathbf{b}_i = \sigma(\mathbf{W}_b \mathbf{e}_{q_i}) \in \mathbb{R}^\mathcal{N}, \\ \mathbf{r}_{q_i} = \text{softmax}(\mathbf{r}_{q_i}) \in \mathbb{R}^\mathcal{N}, 
\end{split}
\end{equation}
where \( \sigma(\cdot) \) is the sigmoid function, and $ \mathbf{W}_1$, $\mathbf{W}_\theta$, $\mathbf{W}_a$, $\mathbf{W}_b $ are learnable weights. 
\paragraph{Training.}  
We train NIRT-Router on a dataset \( \mathcal{D'}_{\text{train}} \) containing tuples \( (q_i, M_j, \mathbf{r}_{q_i}, y_{ij}) \), where \( \mathbf{r}_{q_i} \) is the relevance vector of the query.


The loss function remains the same as in MIRT-Router, using a binary cross-entropy loss (Eq.(\ref{loss})).
\subsection{Routing Decision}
After obtaining the predicted performance $\hat{\mathcal{P}}(q_i, M_j)$, we combined it with the LLM fixed cost $\mathcal{C}(M_j)$ using the score function (Eq.(\ref{score})). The LLM with the highest score is the one to which the query is routed.
\subsection{Warm-Up for Query Cold-Start}
In real-world scenarios, test queries are typically unseen during training, leading to the cold-start problem. Although text-based semantic embeddings somewhat alleviate this issue, we have implemented a further improvement through refining the query vector by incorporating information from similar known queries.
Specifically, given a test query , we update its vector as
\begin{equation}
\mathbf{e}_{q_i} = (1 - \lambda) \cdot \mathbf{e}_{q_i} + \lambda \cdot \mathbf{e}^{\text{warm}}_{q_i},
\end{equation}
where $\mathbf{e}^{\text{warm}}_{q_i}$ is an adjustment vector obtained by averaging the embeddings of its k-nearest neighbors in the training set, identified using a similarity search in the query embedding space.

\section{Experimental Setup}
\label{eval}
This section will provide a detailed explanation of the data construction and setup. As mentioned in Section~\ref{train}, the training set is defined as $\mathcal{D}_{\text{train}} = \{(q_i, M_j, y_{ij}), q_i \in \mathcal{Q}_\text{train}, M_j \in \mathcal{M}, y_{ij} \in [0,1]\}$, and the test set follows a similar structure.

Specifically, we construct interaction data between 12 different types of datasets and 20 different LLMs. For each query, we generate responses from all 20 LLMs. The response quality is then evaluated against the ground truth using the corresponding evaluation metrics described in Table~\ref{tab:datasets}, producing the performance score of $M_j$ on $q_i$, which is $ y_{ij} $.

\subsection{Datasets}
\paragraph{In-distribution (ID).} 
In this scenario, we utilized 8 datasets. For each dataset, we randomly split it into a training set (70\%) and a test set (30\%). All training sets were combined to form the overall training query set \( \mathcal{Q}_{\text{train}} \) for learning the router. Similarly, all test sets were combined to form the overall test set \( \mathcal{Q}_{\text{test}} \), which was used to evaluate the router in an in-distribution scenario. Since the training and test query sets were partitioned before interacting with the LLMs, all queries in the test set were unseen.
The 8 datasets are as follows: \textbf{(1) MMLU} \cite{hendrycks2021measuringmassivemultitasklanguage}: A benchmark including 57 tasks across diverse domains. \textbf{(2) CMMLU} \cite{li2024cmmlumeasuringmassivemultitask}: A Chinese multitask evaluation benchmark covering 67 tasks. \textbf{(3) ACLUE} \cite{zhang2023largelanguagemodelcomprehend}: An ancient Chinese language understanding benchmark. \textbf{(4) ARC\_C} \cite{clark2018think}: A dataset designed to measure advanced reasoning capabilities. \textbf{(5) Hotpot\_QA} \cite{yang2018hotpotqa}: A dataset that requires multi-hop reasoning across documents. \textbf{(6) SQUAD} \cite{rajpurkar2018know}: A reading comprehension dataset consisting of questions posed by crowdworkers. \textbf{(7)MATH} \cite{hendrycks2021measuring}: A dataset of lots of challenging competition mathematics problems. \textbf{(8) MBPP} \cite{austin2021program}: A dataset of programming tasks.

\paragraph{Out-of-distribution (OOD).}
We also evaluate the trained router on 4 OOD datasets, which are:
\textbf{(1) CEVAL} \cite{huang2024c}: A Chinese dataset spanning 52 disciplines. \textbf{(2) Commonsense\_QA} \cite{talmor2018commonsenseqa}: tests commonsense reasoning. \textbf{(3) GSM8K} \cite{cobbe2021training}: contains grade-school math word problems. \textbf{(4) HumanEval:} evaluates code generation capabilities.
\subsection{Candidate LLMs}

As mentioned in Section~\ref{intro}, there are various types of LLMs available. Here, we select 20 representative models (listed in Table~\ref{tab:llms}) as candidate LLMs.

\subsection{Baselines}

We compare our proposed methods (MIRT-Router and NIRT-Router) with \textbf{Small LLM} (i.e., Ministral-8B-Instruct-2410) and \textbf{Large LLM} (i.e., GPT-4o), which always route queries to the small and large models, respectively, as well as with some recent representative works, as follows:
\textbf{HybridLLM}\cite{ding2024hybrid} and \textbf{RouteLLM}\cite{ong2024routellm} use DeBERTa-v3~\cite{he2023debertav3improvingdebertausing} and Matrix Factorization (MF) as binary classifiers to select the better-performing model between a pair of large and small models, respectively. We define the small model as Ministral-8B-Instruct-2410 and the large model as GPT-4o for both.
\textbf{RouterBench}\cite{hu2024routerbench} designs multiple routing strategies to respond to user queries and route them to the best LLM from a set of candidates. We adopt the ``Predictive Router'' strategy for our implementation, as other strategies require obtaining responses from each candidate LLM first, which incurs higher costs.

\begin{table*}[]
\caption{Testing results in In-distribution scenario. Performance, Total Cost(\$) and Reward($\times 10^{-2}$) are three metrics mentioned in Section~\ref{metric}. The best results on each setting are highlighted. The bigger $\alpha$ is, the more requirements for performance. $\downarrow$ indicates the lower, the better. $\uparrow$ indicates the higher, the better. }
\renewcommand{\arraystretch}{1.1}
\centering
\scalebox{0.65}{
\begin{tabular}{c|ccc|ccc|ccc}
\hline
\textbf{Setting} & \multicolumn{3}{c|}{$\bm{\alpha = 0.8}$, $\bm{\beta = 0.2}$} & \multicolumn{3}{c|}{$\bm{\alpha = 0.5}$, $\bm{\beta = 0.5}$} & \multicolumn{3}{c}{$\bm{\alpha = 0.2}$, $\bm{\beta = 0.8}$} \\ \cline{2-10} \cline{2-10}
& \textbf{Performance↑} & \textbf{Total Cost↓} & \textbf{Reward↑} & \textbf{Performance↑} & \textbf{Total Cost↓} & \textbf{Reward↑} & \textbf{Performance↑} & \textbf{Total Cost↓} & \textbf{Reward↑} \\ \hline
Small LLM &  48.70\%&  \textbf{0.31} &   38.48&  48.70\%&  \textbf{0.31} &  23.14&  48.70\%&  \textbf{0.31} &  7.81\\ 
Large LLM &  77.53\%&  12.93 &  42.02&  77.53\%&  12.93&  -11.24&  77.53\%&  12.93 &  -64.49\\
\cline{1-10}
HybridLLM &  54.37\%&  1.98 &  40.43&  52.42\%&  1.54 &  20.27&  56.65\%&  2.78 &  -5.84\\
RouteLLM &  77.25\%&  12.80 &  42.00&  73.59\%&  11.15 &  -6.32&  66.24\%&  7.51 &  -33.17\\
RouterBench &  80.01\%&  1.15 &  62.23&  79.48\%&  0.53 &  37.67&  78.36\%&  0.37 &  13.35\\
\cline{1-10}
\textbf{MIRT-Router} & 80.67\%&  0.42& \textbf{63.89} &  \textbf{80.65\%}&  0.42 & \textbf{38.69} &  \textbf{80.03\%}&  0.39 &  \textbf{13.59}\\
\textbf{NIRT-Router} & \textbf{80.69\%} & 0.55  & 63.70 & 80.41\% & 0.43 & 38.53 & 79.37\% & 0.41 & 13.37  \\
\hline 
\end{tabular}
}
\label{table:id}
\end{table*}

\begin{table*}[]
\caption{Testing results in Out-of-distribution scenario.}
\renewcommand{\arraystretch}{1.1}
\centering
\scalebox{0.65}{
\begin{tabular}{c|ccc|ccc|ccc}
\hline
\textbf{Setting} & \multicolumn{3}{c|}{$\bm{\alpha = 0.8}$, $\bm{\beta = 0.2}$} & \multicolumn{3}{c|}{$\bm{\alpha = 0.5}$, $\bm{\beta = 0.5}$} & \multicolumn{3}{c}{$\bm{\alpha = 0.2}$, $\bm{\beta = 0.8}$} \\ \cline{2-10} \cline{2-10}
& \textbf{Performance↑} & \textbf{Total Cost↓} & \textbf{Reward↑} & \textbf{Performance↑} & \textbf{Total Cost↓} & \textbf{Reward↑} & \textbf{Performance↑} & \textbf{Total Cost↓} & \textbf{Reward↑} \\ \hline
Small LLM &  59.83\%&  \textbf{0.11} &  45.67&  59.83\%&  \textbf{0.11} &  28.88&  59.83\%&  \textbf{0.11} &  10.31\\ 
Large LLM &  84.90\%&  5.30 &  47.92&  84.90\%&  5.30 &  -7.55&  84.90\%&  5.30 &  -63.02\\
\cline{1-10}
HybridLLM &  63.34\%&  0.73 & 47.93 &  62.08\%&  0.41 &  27.19&  63.79\%&  0.65 &  2.98\\
RouteLLM &  84.39\%&  5.25 &  47.70&  79.90\%&  4.74 &  -4.83&  75.06\%&  3.48 &  -37.58\\
RouterBench &  85.50\%&  0.26 &  67.42&  85.75\%&  0.16 &  41.37&  84.62\%&  0.12 &  15.09\\
\cline{1-10}
\textbf{MIRT-Router} &  87.12\%&  0.14 &  69.17&  87.12\%&  0.14 &  42.25&  87.18\%&  0.13 &  15.45\\
\textbf{NIRT-Router} & \textbf{87.37\%} & 0.15 & \textbf{69.32} & \textbf{87.24\%} & 0.14 & \textbf{42.30} & \textbf{87.46\%} & 0.13 & \textbf{15.50}\\
\hline 
\end{tabular}
}
\label{table:ood}
\end{table*}

\subsection{Metrics}
\label{metric}
Following GraphRouter \citep{feng2024graphrouter}, we evaluate all routing methods using three metrics:

\paragraph{Performance}
The average response performance across all test queries, which is evaluated against the ground truth (see Section \ref{eval} and Table \ref{tab:datasets}).

\paragraph{Total Cost}
The total expenditure incurred for all test queries (measured in USD). It is computed as:

\begin{small}
\begin{equation}
\begin{split}
\sum_{q \in \mathcal{Q}_\text{test} } [\text{input\_pricing}(M^*(q)) \times  \text{input\_tokens}(q) + \\ \text{output\_pricing}(M^*(q)) \times \text{output\_tokens}(q, M^*(q))].
\end{split}
\end{equation}
\end{small}

\paragraph{Reward}
To unify the measurement, we follow the normalization approach in Section \ref{pd}; the \text{Total Cost} is linearly mapped to the range $[0,1]$. The scaling factor is determined by the maximum observed Total Cost. The final reward function balances performance and cost:

\begin{small}
\begin{equation}
\text{Reward} = \alpha \cdot \text{Performance} - \beta \cdot \text{linear}(\text{Total Cost}).
\end{equation}
\end{small}

\subsection{Implementation Details}
We use \textit{bert-base-uncased}\footnote{https://huggingface.co/google-bert/bert-base-uncased} as embedding model for both queries and LLMs. The k in warm-up mechanism is set to 5. And the Dimension $\mathcal{N}$ of MIRT-Router and NIRT-Router are both set to 25. The router is trained using the Adam optimizer with a learning rate of 0.002 and a batch size of 512. All experiments are run on 1 NVIDIA A100 40GB GPU.

\begin{figure*}[]
  \includegraphics[width=\linewidth]{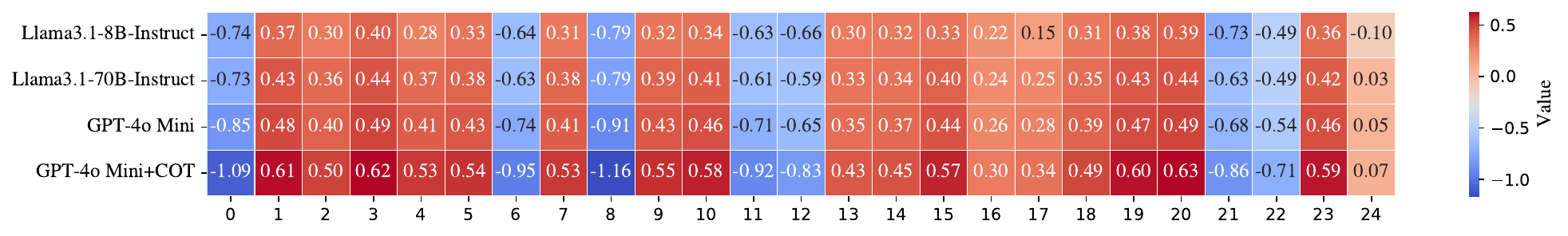} 
  \caption{Four LLMs' ability values across 25 dimensions. From top to bottom, the 4 average ability values are \textbf{0.0257}, \textbf{0.0763}, \textbf{0.0841}, and \textbf{0.0971}. Higher values indicate stronger capability.}
  \label{fig:ma}
\end{figure*}

\section{Experimental Results}
\subsection{Main Results}
\paragraph{In-Distribution Results.} From Table \ref{table:id}, we can observe that in all three different settings, our IRT-Router achieves the highest performance. The average answer accuracy is 3\% higher than when using GPT-4o alone, yet the Total Cost is only 1/30 of GPT-4o’s. When compared to using only small models, our method improves the Performance by 32\%, with a comparable Total Cost. Moreover, under all settings that balance performance and cost (Metric Reward), IRT-Router also achieves the optimal results.
Through comparisons with IRT-Router and RouterBench against other baselines that only use two candidate LLMs, it is clear that multi-LLM routing significantly outperforms binary-LLM routing. This indicates that binary-LLM routing fails to fully exploit the complementary capabilities of different LLMs.
We also observe that while IRT-Router outperforms RouterBench, especially in the case of $\alpha=0.8$ (performance priority), where it not only provides better performance but also costs only half of RouterBench. But RouterBench adheres more strictly to variations in $\alpha$. We think this is related to the measurement or definition of the LLM’s fixed cost, $\mathcal{C}(M_j)$.
Finally, we find that the performance of MIRT-Router is very similar to NIRT-Router, with MIRT-Router slightly outperforming NIRT-Router in the ID scenario.

\paragraph{Out-of-Distribution Results.}
In the OOD scenario, IRT-Router also demonstrates strong performance, achieving the highest Reward. Its Performance is 2\% higher than the best-performing baseline. Notably, NIRT-Router outperforms MIRT-Router in this scenario, suggesting that NIRT-Router may have slightly better generalization ability compared to MIRT-Router. We believe this is due to the more complex network structure of NIRT-Router.

\begin{figure}[]
\centering
  \includegraphics[width=\columnwidth]{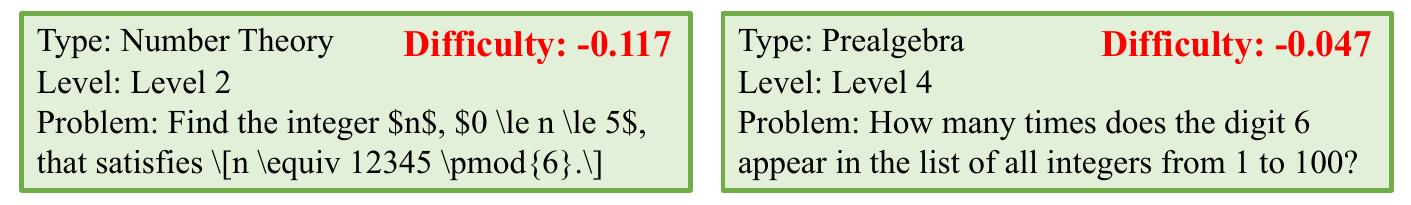}
  
  \caption {Two queries from the MATH dataset. The higher the values of Level and Difficulty, the more challenging the query.}
  \label{fig:md}
  
\end{figure}

\subsection{Interpretability of IRT-Router}
\paragraph{LLM Ability}
We select two pairs of models from the same series (Llama3.1-8B-Instruct vs Llama3.1-70B-Instruct, GPT-4o Mini vs GPT-4o Mini+COT) for ability comparison. As shown in Figure \ref{fig:ma}, the ability values for all four models are obtained from the trained MIRT-Router. It can be seen that Llama3.1-70B-Instruct is either equal to or surpasses Llama3.1-8B-Instruct in each dimension, which aligns with the expected relationship between larger and smaller models, and is consistent with their average performance on ID test set (71\% and 32\%). The other pair involves the COT-enhanced and non-enhanced versions. It is evident that the average ability value of GPT-4o Mini+COT is higher than that of GPT-4o Mini.

\paragraph{Query Difficulty}
We randomly select 2 queries from the MATH dataset, each labeled with a question level. As shown in Figure \ref{fig:md}, the query difficulty derived from MIRT-Router matches the level labels.

\paragraph{Routing Analysis}
The ability values of 20 LLMs obtained from MIRT-Router are ranked, with the top 10 being DeepSeek-Chat (81\%), DeepSeek-Coder (81\%), Gemini-1.5-Flash (73\%), GLM-4-Plus (79\%), GPT-4o (78\%), GPT-4o Mini (71\%), GPT-4o Mini+COT (72\%), Llama3.1-405B-Instruct (78\%), Qwen2.5-32B-Instruct-GPTQ-Int4 (78\%), Qwen2.5-72B-Instruct (80\%), where $(\cdot)$ represents their average performance on the ID training set. In fact, the abilities exhibited by these 10 models are not significantly different, but there is a large gap compared to models like QwQ-32B-Preview (60\%) and Llama3.1-8B-Instruct (32\%). Among the top 10, the lowest cost models are DeepSeek-Chat (\$0.28/M), DeepSeek-Coder (\$0.28/M) and Qwen2.5-32B-Instruct-GPTQ-Int4 (\$0.2/M). We sort all queries of the ID test set by their difficulty (obtained from MIRT-Router) and select the top 30\% and bottom 30\% of queries. We then observe the actual assignment of queries under the setting $\alpha=0.8$. We find that, in the top 30\%, 80\% of the queries are routed to DeepSeek-Chat. In the bottom 30\%, 99\% of the queries are routed to Qwen2.5-32B-Instruct-GPTQ-Int4. This shows that queries with higher difficulty tend to be routed to models with stronger abilities, while queries with lower difficulty are routed to slightly weaker but sufficiently capable and more cost-effective models. It demonstrates the effectiveness and rationality of the IRT-Router.

\paragraph{Routing Accuracy}
To further evaluate the effectiveness of IRT-Router, we additionally assess how accurately MIRT-Router routes unseen queries to the Top-$k$ most optimal LLMs.
The routing accuracy under both in-distribution and out-of-distribution scenarios (with a total of 20 candidate LLMs and $\alpha=0.8$) is reported in Table~\ref{tab:topk_routing_accuracy}.
\begin{table}[]
\centering
\caption{Top-$k$ routing accuracy of MIRT-Router.}
\label{tab:topk_routing_accuracy}
\scalebox{0.8}{
\begin{tabular}{|c|c|c|c|c|c|}
\hline
\textbf{Scenario} & \textbf{Top-1} & \textbf{Top-2} & \textbf{Top-3} & \textbf{Top-4} & \textbf{Top-5} \\
\hline
ID & 2.72\% & 9.88\% & 32.51\% & 38.78\% & 47.85\% \\
OOD & 2.15\% & 7.50\% & 27.29\% & 31.94\% & 39.47\% \\
\hline
\end{tabular}
}
\end{table}
Although the Top-1 accuracy is relatively low, this is mainly due to two factors. First, the routing objective considers both performance and cost. With many candidate LLMs available, several models (e.g., DeepSeek-Coder and DeepSeek-Chat) perform similarly. And even smaller and larger models may get comparable scores, making limiting to just the Top-1 model less meaningful. Second, the current IRT-Router is intentionally lightweight, with a very small number of trainable parameters. With more high-quality training data and refined constraints in the future, allowing IRT-Router to evolve and train continually, we expect its accuracy to improve further.

\subsection{Generalization Ability}
\paragraph{New LLM}
We also conduct experiments on the new LLM, Claude 3.5 Haiku 20241022. Here, we focus on evaluating how different routers perform in predicting the quality of the new LLM's responses on the ID test set. We select four metrics: regression (MAE, RMSE) and classification (AUC, ACC). As shown in Table \ref{table:new}, RouterBench has very low prediction accuracy, almost random, whereas MIRT-Router and NIRT-Router perform much better. Nevertheless, the current IRT-Router still shows limited generalization to unseen LLMs, with an ACC of 0.67, indicating significant room for improvement. We believe improving this aspect is also an important future direction, such as leveraging few-shot learning or similarity to warm up LLM cold-start.

\begin{table}[]
\caption{Results in ID scenario when $\alpha=0.8$.}
\renewcommand{\arraystretch}{1.1}
\centering
\scalebox{0.75}{
\begin{tabular}{c|cccc}
\hline
 & \textbf{RMSE↓} & \textbf{MAE↓} & \textbf{AUC↑} & \textbf{ACC↑} \\ \hline
RouterBench &  0.62&  0.55 & 0.50 &  0.34\\
MIRT-Router &  0.45&  0.43 &  0.62&  0.67\\
NIRT-Router &  \textbf{0.45}&  \textbf{0.42} &  \textbf{0.62}&  \textbf{0.68}\\
\hline 
\end{tabular}
}
\label{table:new}
\end{table}

\paragraph{Warm up Query Cold-Start}

\begin{figure}[]
\centering
  \includegraphics[width=0.48\columnwidth]{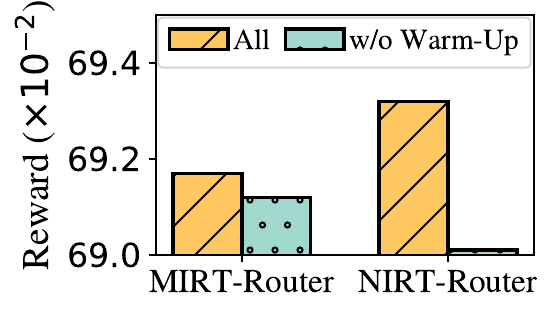}
  \includegraphics[width=0.48\columnwidth]{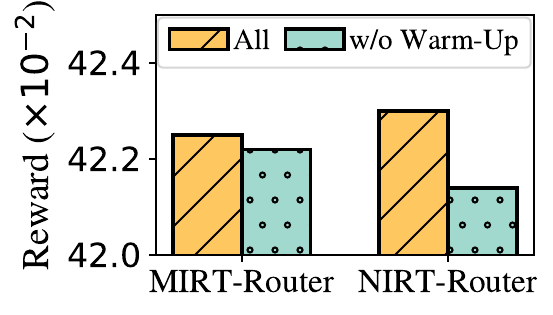}
  \caption {Reward of IRT-Router (All and w/o Warm-up) When $\alpha=0.8$ (left) and $\alpha=0.5$ (right).}
  \label{fig:wo}
\end{figure}

We conduct experiments to study the effectiveness of warm-up for the query cold-start. Figure \ref{fig:wo} and \ref{fig:a2} shows the Reward of MIRT-Router and NIRT-Router (with or without warm-up) on the OOD test set when $\alpha=0.8$, $\alpha=0.5$ and $\alpha=0.2$. We find that when the warm-up module is removed, all rewards decrease, with this effect being more pronounced in NIRT-Router. This indicates that the warm-up mechanism has a more significant impact on improving the performance of NIRT-Router.

\section{Conclusion}
In this work, we introduced IRT-Router, an interpretable and effective LLM router based on Item Response Theory. Our method effectively achieved higher performance and lower cost by selecting the most suitable LLM for a given query. Through extensive experiments on 20 LLMs and 12 datasets, we found that IRT-Router outperformed multiple baselines, demonstrating the superiority of our method. Additionally, our warm-up mechanism for query cold-starts enhanced generalization to unseen queries.

\section*{Limitations}
The datasets currently used are common benchmark datasets with ground truth labels. However, the queries in these datasets are relatively short and do not cover the wide variety encountered in real-world usage. We recognize this as a common challenge in the LLM Router field. Moving forward, it would be valuable to continuously gather dynamic data based on human preferences \cite{zheng2023judging, chiang2024chatbot}, which could better reflect real-world query distributions.

Additionally, our router appears insufficiently sensitive to changes in \( \alpha \), suggesting that a more refined measurement approach is needed, such as increasing the value of the LLM's fixed cost \( \mathcal{C}(M_j) \).

Moreover, we have not imposed additional constraints on the relationship between query attributes and LLM abilities. For instance, if we assume that larger models generally exhibit higher average ability levels than smaller models, we could introduce an ordering constraint on the average values of their learned ability vectors during training. This constraint could guide the optimization process, accelerate convergence, and lead to more reasonable and accurate training outcomes.

\section*{Acknowledgments}
This research was partially supported by the National Science and Technology Major Project (No.2022ZD0117103), the National Natural Science Foundation of China (Grants No.62477044), the Fundamental Research Funds for the Central Universities (No.WK2150110038), and CCF-NetEase ThunderFire Innovation Research Funding (NO. CCF-Netease 202306). Zhenya Huang gratefully acknowledges the support of the Young Elite Scientists Sponsorship Program by CAST (No. 2024QNRC001).


\bibliography{main}

\clearpage

\appendix
\label{sec:appendix}

\section{Experimental details}
\subsection{LLM Profile}
Table \ref{tab:profile} presents the profiles used for LLM embeddings. Here, we obtained each LLM’s profile by utilizing ChatGPT’s web search mode with prompts. Of course, the final results were manually corrected.

\subsection{Cadidate LLMs and their Pricing}
 Cadidate LLMs are categorized into four groups (exists overlapping):

\noindent\raisebox{0.25ex}{\small{$\bullet$}}\normalsize\; \textbf{API-based Models:} Large models accessible via API calls. For closed-source models (e.g., GPT-4o), pricing is based on official API rates, while for open-source models (e.g., Llama3.1-405B-Instruct), we refer to pricing from Together AI \footnote{https://www.together.ai/}.

\noindent\raisebox{0.25ex}{\small{$\bullet$}}\normalsize\; \textbf{Deployable Models:} Smaller models, such as 7B, 8B, or quantized versions(e.g., Ministral-8B-Instruct-2410, Qwen2.5-32B-Instruct-GPTQ-Int4), which can be deployed locally. We standardize inference on a single A100 40GB GPU, with pricing set at \$0.2 per million output tokens, following Together AI.

\noindent\raisebox{0.25ex}{\small{$\bullet$}}\normalsize\; \textbf{Specialized Models:} Models tailored for specific proprietary tasks (e.g., QwQ-32B-Preview, DeepSeek-Coder).

\noindent\raisebox{0.25ex}{\small{$\bullet$}}\normalsize\; \textbf{Enhanced Models:} Models such as GPT-4o Mini  with Chain-of-Thought (CoT) prompting. It shares the same pricing as GPT-4o Mini  but differs in prompting strategies.

The first 20 LLMs in Table \ref{tab:llms} are used in the main experiments, and the last LLM is for validating the generalization of IRT-Router on new LLMs.

\begin{figure}[]
\centering
  \includegraphics[width=0.48\columnwidth]{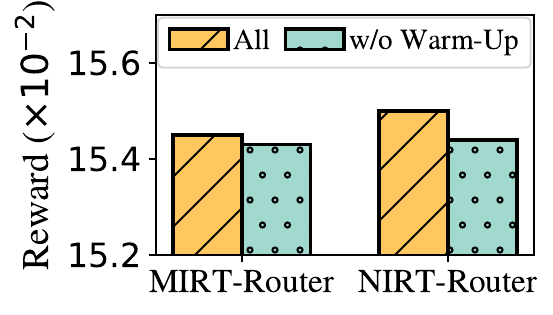}
  \caption{Reward of IRT-Router (All and w/o Warm-up) When $\alpha=0.2$.}
  \label{fig:a2}
\centering
\end{figure}

\begin{table}[h]
  \centering
  \scalebox{0.7}{
  \begin{tabular}{lcc}
    \hline
    \textbf{LLM} & \textbf{Iutput \$/1M} & \textbf{Output \$/1M}\\
    \hline
    DeepSeek-Chat   &   0.14        &   0.28\\
    DeepSeek-Coder  &   0.14       &    0.28\\
    Gemini-1.5-Flash  &   0.075     &      0.3\\
    GLM-4-Air     & 0.137           & 0.137\\
    GLM-4-Flash      & 0.0137            & 0.0137\\
    GLM-4-Plus     & 6.85           & 6.85\\
    GPT-4o     & 2.5           & 10\\
    GPT-4o Mini     & 0.15           & 0.6\\
    GPT-4o Mini+COT     & 0.15           & 0.6\\
    Llama3.1-8B-Instruct     & 0.1           & 0.2\\
    Llama3.1-70B-Instruct     & 0.792           & 0.792\\
    Llama3.1-405B-Instruct       & 3.15            & 3.15\\
    Ministral-8B-Instruct-2410      & 0.1           & 0.2\\
    Mistral-7B-Instruct-v0.2      & 0.1           & 0.2\\
    Mixtral-8x7B-Instruct     & 0.54           & 0.54\\
    Qwen2.5-32B-Instruct-GPTQ-Int4     & 0.1           & 0.2\\
    Qwen2.5-7B-Instruct     & 0.1           & 0.2\\
    Qwen2.5-72B-Instruct     & 1.08           & 1.08\\
    Qwen2.5-Math-7B-Instruct      & 0.1            & 0.2\\
    QwQ-32B-Preview     & 1.2           & 1.2\\
    \hline
    Claude 3.5 Haiku 20241022 (New)    & 0.8           & 4\\
    \hline
  \end{tabular}
  }
  \caption{Cadidate LLMs and their pricing. In practice, the pricing can be adjusted as needed at any time.}
  \label{tab:llms}
\end{table}

\begin{table}[]
\caption{Embedding Models.}
\renewcommand{\arraystretch}{1.1}
\centering
\scalebox{0.8}{
\begin{tabular}{c|cc}
\hline
 & \textbf{Dimension} & \textbf{Pricing} \\ \hline
bge-m3 &  1536& free \\
text-embedding-3-small &  1024& \$0.02 / 1M\\
bert-base-uncased &  768&free \\
zhipu-embedding-3 &  512& \$0.0685 / 1M\\
\hline 
\end{tabular}
}
\label{table:embm}
\end{table}

\begin{table*}[h]
  \centering
  \scalebox{0.95}{
  \begin{tabular}{lcccc}
    \multicolumn{5}{c}{\textbf{In-distribution}} \\
    \hline
    \textbf{Dataset} & \textbf{Type} & \textbf{Evaluation Metric} & \textbf{Train Size} & \textbf{Test Size} \\
    \hline
    ACLUE & Ancient Chinese & accuracy & 1400 & 600 \\
    ARC\_C & Reasoning & accuracy & 1400 & 600 \\
    CMMLU & Chinese Multitask & accuracy & 7000 & 3000 \\
    Hotpot\_QA & Multi-Hop & EM & 1400 & 600 \\
    MATH & Math & accuracy & 1400 & 600 \\
    MBPP & Code & pass@1 & 630 & 270 \\
    MMLU & Multitask & accuracy & 9800 & 4200 \\
    SQUAD & Reading Comprehension & f1 & 1400 & 600 \\
    \hline
    & & & & \\ 
    \multicolumn{5}{c}{\textbf{Out-of-distribution}} \\
    \hline
    \textbf{Dataset} & \textbf{Task type} & \textbf{Evaluation Metric} & \textbf{Train Size} & \textbf{Test Size} \\
    \hline
    CEVAL & Chinese Multitask & accuracy & - & 1000 \\
    Commonsense\_QA & Commonsense Reasoning & accuracy & - & 1000 \\
    GSM8K & Math & accuracy & - & 1000 \\
    HumanEval & Code & pass@1 & - & 160 \\
    \hline
  \end{tabular}
  }
  \caption{Datasets Details.}
  \label{tab:datasets}
\end{table*}

\subsection{Datasets Details}
Table \ref{tab:datasets} shows the details of datasets. So the overall size of $\mathcal{Q}_{\text{train}}$ is $24430$. And the size of $\mathcal{D}_{\text{train}}$ for traning the router is $24430 \times 20 = 488600$.

\section{Sensitivity Analysis}
\paragraph{Embedding Models}
In the previous experiments, we use the \textit{bert-base-uncased} as embedding model for both queries and LLMs. However, there are now many available embedding models, each with different dimensions and pricing. We cannot guarantee that BERT is the optimal choice for every scene. Therefore, we conducted experiments to evaluate the impact of four commonly used embedding models (listed in Table \ref{table:embm}) on the router.

We select four common embedding models: two are provided by large model vendors and require a fee (OpenAI's text-embedding-3-small and Zhipu AI's embedding-3), while the other two are widely used pre-trained language models that can be deployed independently (bge-m3 and bert-base-uncased). The embedding output dimensions for these models are also shown in the Table \ref{table:embm}.

For models that are not free, we also included the embedding cost in the Total Cost calculation. Our findings reveal that while paid embedding models generally offer higher performance, their associated costs are also higher. When considering the trade-off between performance and cost, BERT strikes a favorable balance, yielding relatively higher rewards in our experimental setup.

\begin{table}[]
\caption{Results of MIRT-Router on ID test set when $\alpha=0.8$.}
\renewcommand{\arraystretch}{1.1}
\centering
\scalebox{0.68}{
\begin{tabular}{c|ccc}
\hline
 & \textbf{Performance↑} & \textbf{Total Cost↓} & \textbf{Reward↑} \\ \hline
bge-m3 &  80.65\% & 0.47 &  63.47\\
text-embedding-3-small &  \textbf{80.99\%} &  0.86&  63.53\\
bert-base-uncased &  80.67\% &  \textbf{0.42}&  \textbf{63.89}\\
zhipu-embedding-3 &  80.78\% &  0.59&  63.73\\
\hline 
\end{tabular}
}
\label{table:emb}
\end{table}

\begin{figure}[]
\centering
  \includegraphics[width=0.8\columnwidth]{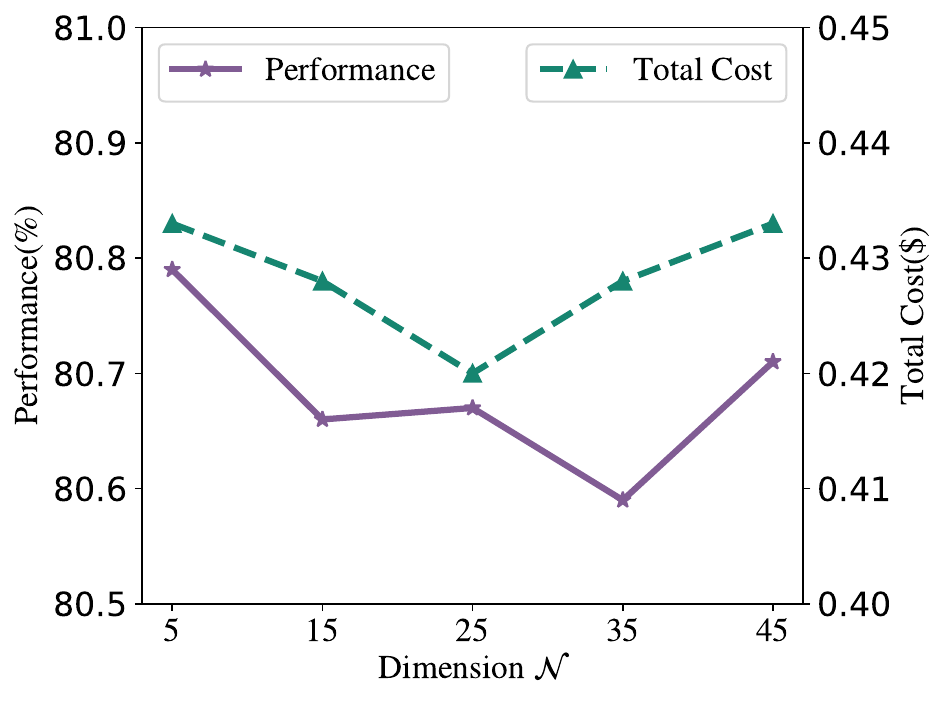}
  \caption {Results of MIRT-Router on ID test set when $\alpha=0.8$ with different dimension $\mathcal{N}$.}
  \label{fig:n}
\end{figure}

\paragraph{Dimension $\mathcal{N}$}
The dimension $\mathcal{N}$ represents the number of dimensions used for modeling the abilities. A larger $\mathcal{N}$ indicates a finer granularity in the ability modeling. But how does $\mathcal{N}$ affect the router's performance? We experiment with five different values of $\mathcal{N}$ and obtain the corresponding results. Figure \ref{fig:n} shows that the Performance fluctuates with changes in $\mathcal{N}$, with no consistent upward or downward trend. On the other hand, the Total Cost tends to be higher when $\mathcal{N}$ is either very small or very large, and it reaches its minimum when $\mathcal{N} = 25$. 

\paragraph{Cold-Start Parameter $\lambda$}
We also conduct ablation studies on the cold-start parameter $\lambda$ under both ID and OOD scenarios. As shown in Table~\ref{tab:lambda}, we vary $\lambda$ from 0 to 0.4 and report the performance, total cost, and reward. For the ID scenario, different $\lambda$ values yield very similar results, with $\lambda=0.2$ or $0.3$ achieving slightly higher reward. In contrast, for the OOD scenario, where cold-start issues are more pronounced, larger $\lambda$ values (e.g., $0.3$ or $0.4$) consistently improve reward. This aligns with intuition, as cold-start issues are more severe in OOD settings.

\begin{table}[]
\centering
\small
\setlength{\tabcolsep}{3.5pt}
\caption{Impact of cold-start parameter $\lambda$ on MIRT-Router under ID and OOD scenarios.}
\label{tab:lambda}
\scalebox{0.82}{
\begin{tabular}{c|ccc|ccc}
\toprule
\multirow{2}{*}{$\lambda$} & \multicolumn{3}{c|}{ID} & \multicolumn{3}{c}{OOD} \\
 & Perf. $\uparrow$ & Total Cost $\downarrow$ & Reward $\uparrow$ & Perf. $\uparrow$ & Total Cost $\downarrow$ & Reward $\uparrow$ \\
\midrule
0   & 80.66 & 0.42 & 63.88 & 87.05 & 0.14 & 69.12 \\
0.1 & 80.66 & 0.42 & 63.88 & 87.08 & 0.14 & 69.14 \\
0.2 & 80.67 & 0.42 & 63.89 & 87.08 & 0.14 & 69.14 \\
0.3 & 80.67 & 0.42 & 63.89 & 87.12 & 0.14 & 69.17 \\
0.4 & 80.64 & 0.42 & 63.86 & 87.12 & 0.14 & 69.17 \\
\bottomrule
\end{tabular}
}
\end{table}





\section{Interpretability of NIRT-Router}
\label{appendix:nirt}
\subsection{Predefined Abilities}
\label{appendix:abilities}
We first predefine the actual meanings corresponding to the \( \mathcal{N} \) dimensions of the ability vector \( \boldsymbol{\theta}_{M_j} \in \mathbb{R}^\mathcal{N} \) in NIRT-Router. In our experiments, \( \mathcal{N} = 25 \). Therefore, we refer to both LLM evaluation and human comment to define the 25 specific abilities corresponding to these 25 dimensions of $\boldsymbol{\theta}_{M_j}$. Notably, these definitions can be adjusted as needed.

The 25 predefined specific ability are as follows:

\raisebox{0.25ex}{\small{$\bullet$}}\normalsize\; 0: Reasoning

\raisebox{0.25ex}{\small{$\bullet$}}\normalsize\; 1: Understanding

\raisebox{0.25ex}{\small{$\bullet$}}\normalsize\; 2: Generation

\raisebox{0.25ex}{\small{$\bullet$}}\normalsize\; 3: Information retrieval

\raisebox{0.25ex}{\small{$\bullet$}}\normalsize\; 4: Multidisciplinary knowledge

\raisebox{0.25ex}{\small{$\bullet$}}\normalsize\; 5: Emotion understanding and expression

\raisebox{0.25ex}{\small{$\bullet$}}\normalsize\; 6: Adaptability and robustness

\raisebox{0.25ex}{\small{$\bullet$}}\normalsize\; 7: Interactivity

\raisebox{0.25ex}{\small{$\bullet$}}\normalsize\; 8: Ethical and moral consideration

\raisebox{0.25ex}{\small{$\bullet$}}\normalsize\; 9: Mathematical calculation

\raisebox{0.25ex}{\small{$\bullet$}}\normalsize\; 10: Data analysis

\raisebox{0.25ex}{\small{$\bullet$}}\normalsize\; 11: Symbolic processing

\raisebox{0.25ex}{\small{$\bullet$}}\normalsize\; 12: Geometric and spatial reasoning

\raisebox{0.25ex}{\small{$\bullet$}}\normalsize\; 13: Programming and algorithms

\raisebox{0.25ex}{\small{$\bullet$}}\normalsize\; 14: Scientific knowledge application

\raisebox{0.25ex}{\small{$\bullet$}}\normalsize\; 15: Technical documentation understanding

\raisebox{0.25ex}{\small{$\bullet$}}\normalsize\; 16: Current affairs and common knowledge

\raisebox{0.25ex}{\small{$\bullet$}}\normalsize\; 17: Cultural understanding

\raisebox{0.25ex}{\small{$\bullet$}}\normalsize\; 18: Language conversion

\raisebox{0.25ex}{\small{$\bullet$}}\normalsize\; 19: Music and art understanding

\raisebox{0.25ex}{\small{$\bullet$}}\normalsize\; 20: Editing and proofreading

\raisebox{0.25ex}{\small{$\bullet$}}\normalsize\; 21: Prediction and hypothesis testing

\raisebox{0.25ex}{\small{$\bullet$}}\normalsize\; 22: Inference

\raisebox{0.25ex}{\small{$\bullet$}}\normalsize\; 23: Decision support

\raisebox{0.25ex}{\small{$\bullet$}}\normalsize\; 24: Content summarization

\subsection{Prompt for Getting Relevance Vector}
\label{appendix:prompt}
\begin{figure}[h]
\centering
  \includegraphics[width=\columnwidth]{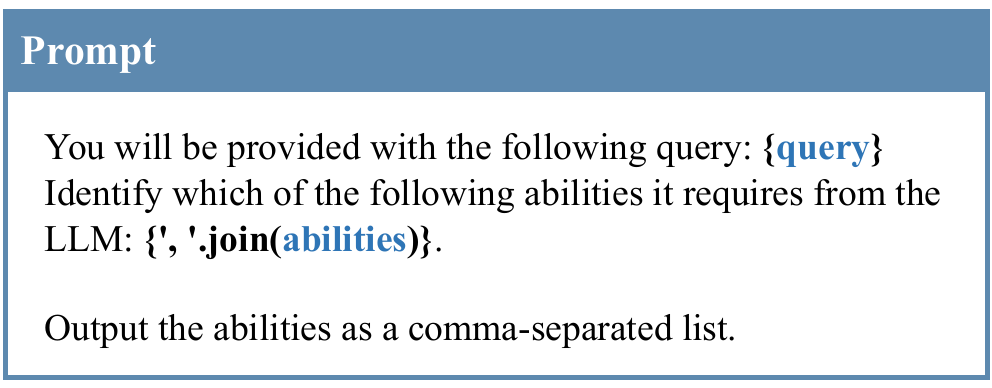}
\centering
\caption{Prompt for getting relevance vector.}
\label{fig:prompt}
\end{figure}

In Section \ref{rele}, we have mentioned that we use a LLM to identify the relevant abilities by considering the abilities required by 5 sample questions within the cluster. Specifically, we use GPT-4o Mini to complete the simple task. And the prompt is shown in Figure \ref{fig:prompt}.

\subsection{LLM Ability Visualization}
\begin{figure}[t]
\centering
  \includegraphics[width=0.9\columnwidth]{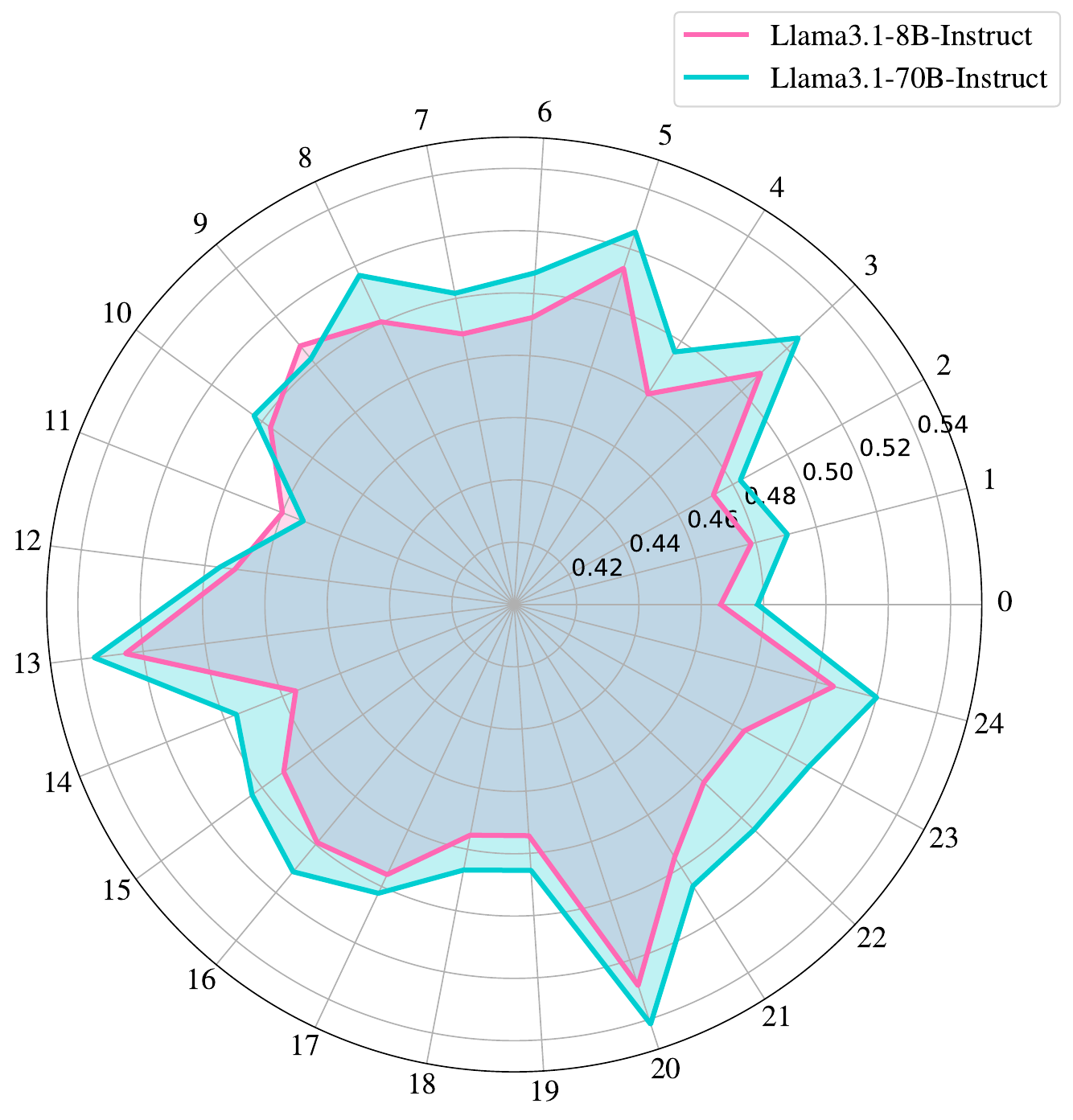} 
  \includegraphics[width=0.9\columnwidth]{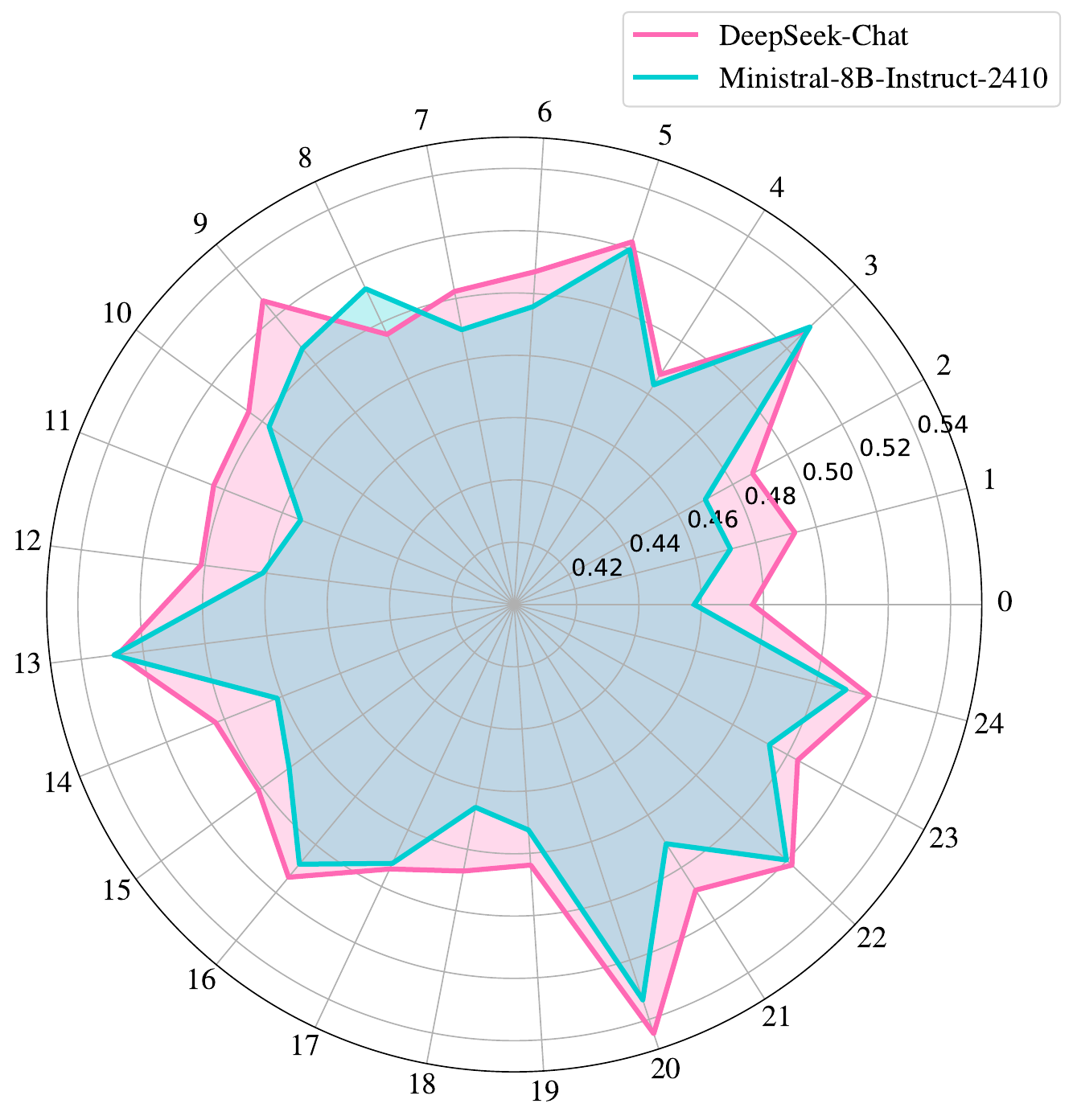}
  \caption {Four LLMs’ ability values across 25 dimensions obtained from the trained NIRT-Router.}
  \label{fig:nirt}
\end{figure}

Figure \ref{fig:nirt} shows four LLMs' ability values across 25 dimensions obtained from the trained NIRT-Router. From the comparison between Llama3.1-8B-Instruct and 70B, the larger model outperforms the smaller one in almost every dimension. The gap is even more pronounced when comparing DeepSeek-Chat and Ministral-8B-Instruct-2410, especially in dimensions 0: Reasoning, 1: Understanding, 9: Mathematical calculation and 11: Symbolic processing, where the larger model significantly surpasses the smaller one. However, in dimension 8: Ethical and moral consideration, the larger model unexpectedly under-performs the smaller one. We speculate that this could be due to insufficient training data in this specific aspect, leading to inadequate learning. Another possible explanation is that the larger model’s broader knowledge base makes it more prone to generating unbounded content.


\begin{table*}
    \centering
    \scalebox{0.6}{
    \begin{tabular}{|c|p{18cm}|} 
    \hline
    \multicolumn{1}{|c|}{\textbf{LLM}} & \multicolumn{1}{c|}{\textbf{Profile}} \\
    \hline
     DeepSeek-Chat& DeepSeek-Chat: Released by DeepSeek AI (China) in December 2024, it is an advanced conversational AI model designed to facilitate human-like interactions. Leveraging the architecture of DeepSeek V3, a 685 billion parameter model, DeepSeek-Chat offers versatile and cost-efficient performance, outperforming models like GPT-4o and Llama 3.1. It is trained on a diverse dataset, enabling it to understand and generate human-like text across various domains.\\
     \hline
     DeepSeek-Coder& DeepSeek-Coder: Released by DeepSeek AI (China) in June 2024, it is an open-source series of code language models designed to enhance code intelligence in software development. The models range from 1.3 billion to 33 billion parameters and are trained from scratch on 2 trillion tokens, with a composition of 87\% code and 13\% natural language in both English and Chinese. They feature a context window of 16,000 tokens and support project-level code completion and infilling. DeepSeek-Coder has demonstrated performance comparable to GPT-4 Turbo in code-specific tasks.\\
     \hline
     Gemini-1.5-Flash& Gemini-1.5-Flash, released by Google in 2024, is a lightweight model optimized for speed and efficiency. It excels in handling high-volume, high-frequency tasks such as summarization, chat applications, and data extraction from long documents and tables, making it suitable for applications requiring quick processing and scalability.\\
     \hline
     GLM-4-Air& GLM-4-Air is released on 2024-06-05 by Zhipu AI, a Chinese AI startup. Best cost-performance model, similar overall performance to GLM-4, with 128k context, fast and affordable. Pre-training corpus consists of multilingual (mostly English and Chinese) documents from a mixture of different sources.\\
     \hline
     GLM-4-Flash& GLM-4-Flash is released for free use in 2024 by Zhipu AI, a Chinese AI startup. Suitable for simple tasks, fastest speed, most affordable version, with 128k context.\\
     \hline
     GLM-4-Plus& GLM-4-Plus is released for use on 2024-08-30 by Zhipu AI, a Chinese AI startup. It's latest base Large Model of Zhipu AI, representing a comprehensive enhancement in language understanding, instruction following, and long-text processing, maintaining an internationally leading level of performance, which demonstrates powerful visual capabilities comparable to OpenAI GPT-4.\\
     \hline
     GPT-4o& GPT-4o is released by OpenAI on May 13, 2024. GPT-4o is a flagship multimodal language model capable of processing text, images, and audio inputs. It features a context window of 128,000 tokens and supports up to 16,000 output tokens per request. The model’s training data includes diverse sources up to October 2023, enhancing its language understanding and generation capabilities.\\
     \hline
     GPT-4o Mini& GPT-4o Mini is introduced by OpenAI on July 18, 2024. GPT-4o Mini is a streamlined, cost-effective version of GPT-4o. It maintains a 128,000-token context window and supports up to 16,000 output tokens per request. Designed for efficiency, GPT-4o Mini offers competitive performance in reasoning, coding, and multimodal tasks, making it suitable for a wide range of applications.\\
     \hline
     GPT-4o Mini+COT& **This version (GPT-4o Mini COT) is enhanced by using Chain of thoughts**. GPT-4o Mini is introduced by OpenAI on July 18, 2024. GPT-4o Mini is a streamlined, cost-effective version of GPT-4o. It maintains a 128,000-token context window and supports up to 16,000 output tokens per request. Designed for efficiency, GPT-4o Mini offers competitive performance in reasoning, coding, and multimodal tasks, making it suitable for a wide range of applications.\\
     \hline
     Llama3.1-8B-Instruct& Llama3.1-8B-Instruct: Released by Meta AI in September 2024, this 8-billion parameter instruction-tuned model is designed to excel in various natural language understanding and generation tasks. It supports a context length of up to 128,000 tokens, enabling the processing of extensive inputs. The model has been fine-tuned on a diverse dataset to enhance its performance in following instructions across multiple domains.\\
     \hline
     Llama3.1-70B-Instruct& Llama3.1-70B-Instruct: Also released by Meta AI in September 2024, this 70-billion parameter instruction-tuned model offers enhanced capabilities in understanding and generating human-like text. With a context length support of up to 128,000 tokens, it is well-suited for complex tasks requiring deep comprehension and extended context handling. The model has been fine-tuned to improve its instruction-following abilities across various applications.\\
     \hline
     Llama3.1-405B-Instruct& Llama3.1-405B-Instruct: This is the largest model in the Llama3.1 series, released by Meta AI in September 2024, featuring 405 billion parameters. It is designed to provide state-of-the-art performance in natural language processing tasks, with a focus on instruction-following capabilities. The model supports a context length of up to 128,000 tokens, making it suitable for highly complex and extended tasks. Fine-tuned on an extensive and diverse dataset, it aims to deliver superior performance across a wide range of applications.\\
     \hline
     Ministral-8B-Instruct-2410& Ministral-8B-Instruct-2410: Released by Mistral AI on October 16, 2024, Ministral-8B-Instruct-2410 is an 8-billion parameter instruction fine-tuned language model. It is designed to significantly outperform existing models of similar size, offering a context window of up to 128,000 tokens. The model has been trained on a diverse dataset, including a substantial proportion of multilingual and code data, enhancing its versatility across various domains.\\
     \hline
     Mistral-7B-Instruct-v0.2& Mistral-7B-Instruct-v0.2: Released by Mistral AI on March 23, 2024, Mistral-7B-Instruct-v0.2 is an instruction-tuned language model with 7 billion parameters. It has been fine-tuned to understand and execute specific instructions effectively, making it suitable for applications such as chatbots, virtual assistants, and task-oriented dialogue systems. The model supports a context window of up to 32,000 tokens, enabling it to process and maintain coherence over longer text sequences.\\
     \hline
     Mixtral-8x7B-Instruct& Mixtral-8x7B-Instruct: Released by Mistral AI in December 2023, Mixtral-8x7B-Instruct is an ensemble model comprising eight 7-billion parameter models, totaling 56 billion parameters. It is designed to deliver enhanced performance across various natural language processing tasks by leveraging the strengths of multiple models. This architecture allows for improved accuracy and robustness in understanding and generating human-like text.\\
     \hline
     Qwen2.5-32B-Instruct-GPTQ-Int4& Qwen2.5-32B-Instruct-GPTQ-Int4: Released by Alibaba Cloud’s Qwen team (China), this is a 32.5 billion parameter instruction-tuned model, quantized to 4-bit precision using GPTQ for efficient inference. It features a context length of up to 131,072 tokens, facilitating long-context understanding and generation.\\
     \hline
     Qwen2.5-7B-Instruct& Qwen2.5-7B-Instruct: This 7.61 billion parameter instruction-tuned model from Alibaba Cloud’s Qwen team (China) is designed for general-purpose language understanding and generation. It incorporates advanced architectural features and supports a context length of up to 131,072 tokens, enabling effective handling of long texts.\\
     \hline
     Qwen2.5-72B-Instruct& Qwen2.5-72B-Instruct: Part of the Qwen2.5 series by Alibaba Cloud (China), this 72 billion parameter instruction-tuned model is tailored for complex language tasks requiring extensive understanding and generation capabilities. It benefits from a large-scale pretraining dataset and advanced architectural design, supporting long-context processing.\\
     \hline
     Qwen2.5-Math-7B-Instruct& Qwen2.5-Math-7B-Instruct: This 7 billion parameter instruction-tuned model from Alibaba Cloud’s Qwen team is designed specifically for mathematical problem-solving and reasoning tasks. It is trained on a specialized mathematical dataset to enhance its proficiency in handling complex mathematical queries.\\
     \hline
     QwQ-32B-Preview& QwQ-32B-Preview is an experimental research model developed by the Qwen Team (China), released in November 2024. It focuses on advancing AI reasoning capabilities, particularly in mathematics and programming. The model has 32.5 billion parameters and a context window of 32,768 tokens. It employs a dense transformer architecture with advanced components such as Rotary Position Embedding (RoPE), SwiGLU activation functions, RMSNorm normalization, and Attention QKV bias.\\
    \hline
    \hline
     Claude 3.5 Haiku 20241022 (New)& Claude 3.5 Haiku 20241022 is a large language model developed by Anthropic, released on October 22, 2024. It features advanced natural language understanding and generation capabilities, supports multiple languages, and is designed to provide efficient and accurate text processing for a wide range of applications, including customer service, content creation, and educational tools.\\
     \hline
    \end{tabular}
    }
    \centering
  \caption{LLMs' profiles. The first 20 LLMs are used in the main experiments, and the last LLM is for validating the generalization of IRT-Router on new LLMs.}
  \label{tab:profile}
\end{table*}


\end{document}